\documentclass[runningheads]{llncs}
\usepackage{graphicx}
\usepackage{subcaption}
\usepackage{graphics}
\usepackage{amsmath}
\usepackage{multirow}
\usepackage{amstext}
\usepackage[textsize=tiny]{todonotes} 
\usepackage[export]{adjustbox}
\usepackage{mathtools}
\usepackage{enumitem}
\usepackage{breqn}
\usepackage{xcolor}
\usepackage{booktabs}
\usepackage{cite}
\usepackage{hyperref}
\usepackage[nameinlink,noabbrev]{cleveref}

\begin{document}
\title{Predicting Gender via Eye Movements\thanks{R.Haria and S. Al Zaidawi contributed equally to the work.}
	}
\titlerunning{Predicting Gender via Eye Movements}
\author{Rishabh Vallabh Varsha Haria\orcidID{0000-0001-8067-3608} \and
Sahar Mahdie Klim Al Zaidawi\orcidID{0000-0003-1504-9264 } \and
Sebastian Maneth\orcidID{0000-0001-8667-5436}}
\authorrunning{R. Haria et al.}
\institute{Department of Informatics in University of Bremen, Germany \\
	\email{haria@uni-bremen.de},
\email{saharmah@uni-bremen.de},
\email{maneth@uni-bremen.de}}
\maketitle             
 In this paper, we report the first stable results on gender prediction
via eye movements. We use a dataset with images of faces as stimuli and
with a large number of 370 participants. 
Stability has two meanings for us: first that we are able to estimate the standard
deviation~(SD) of a single prediction experiment (it is around 4.1\,\%); this is achieved by varying the
number of participants. And second, we are able to provide a mean accuracy with
a very low standard error~(SEM): our accuracy is 65.2\,\%, and the~SEM is 0.80\,\%;
this is achieved through many runs of randomly selecting training and test sets for the prediction. 
Our study shows that two particular classifiers achieve the best accuracies: Random Forests and Logistic Regression. 
Our results reconfirm previous findings
that females are more biased towards the left eyes of the stimuli.


\section{Introduction}
\label{sec:introduction}

A lot of research has been spawned by the advent of new eye tracking devices.
Recordings of eye movements can be used for a variety of purposes;
for instance, to detect particular diseases, to identify persons (biometrics), or to 
study cognitive developments in children. 
Previous studies on using machine learning classifiers for gender prediction via eye movements
have achieved accuracies of 64\,\%~\cite{mercer2012eye} and 70\,\%~\cite{sargezeh2019gender}, respectively. One drawback of these results, however, is their
instability in these two respects: either an unknown or large standard deviation (SD), 
or a low number of participants.
The first study does not mention the SD while the second states an SD of 13.22\,\% and these studies use
52 and 45 users, respectively.
Here, we present the first stable results for gender prediction:
our employed dataset features 370~users. Our results show a mean accuracy of 65.2\,\% based on 50 runs, with a standard error (SEM) of 0.80\,\%, and we estimate the SD of a single prediction experiment to be around 4.1\,\% by varying the number of participants.

Our method segments eye movement trajectories into fixations and saccades by using the I-VT algorithm~\cite{george2016score, mahdie2021extensive, al2020gender}. 
A small number of statistical features are then computed.
We use two separate classifiers of the same kind, one for fixation's and another for saccade's. 
Different ML classification algorithms were examined, and we found that Logistic Regression~\cite{bishop2006pattern_chap4} and Random Forest~\cite{breiman2001random, breiman1996bagging} performed best.
We find that the best accuracy is obtained when we use the Nelder-Mead algorithm~\cite{powell1964efficient} to weight the fixation and the saccade classifiers. We observe increased accuracy when fixation weights are higher.

Another highlight of our results is the fact that our best accuracies are obtained by using only two features:
the maximum angular velocity (this is applied to fixations) and the
saccade ratio (maximum angular velocity divided by the duration; this is applied to saccades). 
We consider such a small number of features an advantage. 

Our main contributions in this paper are: 
\begin{itemize}[noitemsep]
\item We present the first stable results for gender prediction using 50 runs and we achieve an accuracy of 65.2\,\% with SEM of 0.80\,\%. 

\item We compare our approach to the best previously known accuracy of 70\,\% with an SD of 13.22\,\%~\cite{sargezeh2019gender} and we show
that in their setting (45 users and 5 runs) we achieve an accuracy of 77.5\,\% with an SD of only 4.00\,\%.

\item We are also able to confirm previous findings that females 
are biased towards the left eye~\cite{leonards2005idiosyncratic, sammaknejad2017gender, saether2009anchoring} and have a more explorative gaze behavior when compared to males\cite{sargezeh2019gender, al2020gender, li2018biometric}
(the stimuli of our dataset are pictures of human faces) and we show that these differences
are statistically significant.

\end{itemize} 

\section{Proposed System}
\label{Proposed_system}
This section describes our gender prediction architecture including the employed dataset, data preprocessing and segmentation methods, feature extraction, machine learning classifiers, and the accuracy metric used in this study.
\subsection{Dataset}
In our study we use the \emph{gaze on faces} (GOF)\footnote[2]{The GOF dataset can be found here~\cite{FilesUNC97:online}.} dataset.
This dataset~\cite{coutrot2016face} comprises eye movement recordings of 405 participants aged 18--72 years. There were 27 participants whose data was erratic or absent so they were eliminated. After eliminating the participants the age group changed to 20--72 years. The GOF dataset comprised of 370 users with 185 females and 185 males. 
The stimuli employed in this dataset are faces of actors. 
The recordings were produced in 2016~at the Science Museum of London, UK. 
The eye tracker was an EyeLink 1000 eye-tracker running at 250 Hz.
The distance of the seated participants was 57~cm from an LCD screen of 19 inch ($1280 \times 1024$~pixels). 
The stimuli's height and width was $429 \times 720$ pixels. 
Overall, eight actors were used as stimuli consisting of four females and four males. During the start and end of the video clip, the actor gazed towards the bottom of the screen for half a second.

\begin{figure}[!h] 
	\begin{subfigure}[b]{0.5\linewidth}
		\centering
		\fbox{\includegraphics[width=0.6\linewidth,height=0.25\textheight]{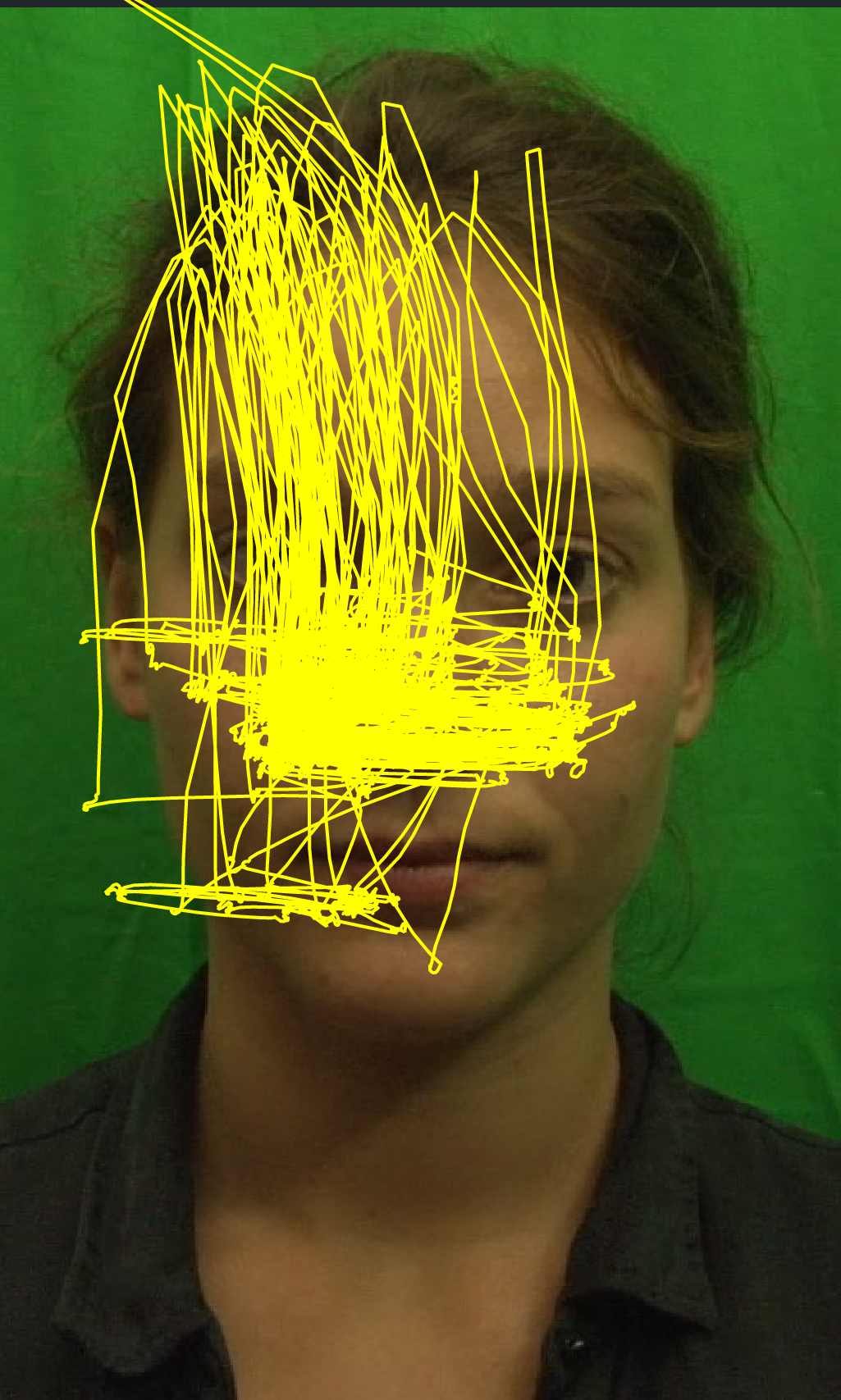}}
		\caption{Female participant} 
		\label{M11} 
	\end{subfigure}
	\begin{subfigure}[b]{0.5\linewidth}
		\centering
		\fbox{\includegraphics[width=.6\linewidth,height=0.25\textheight]{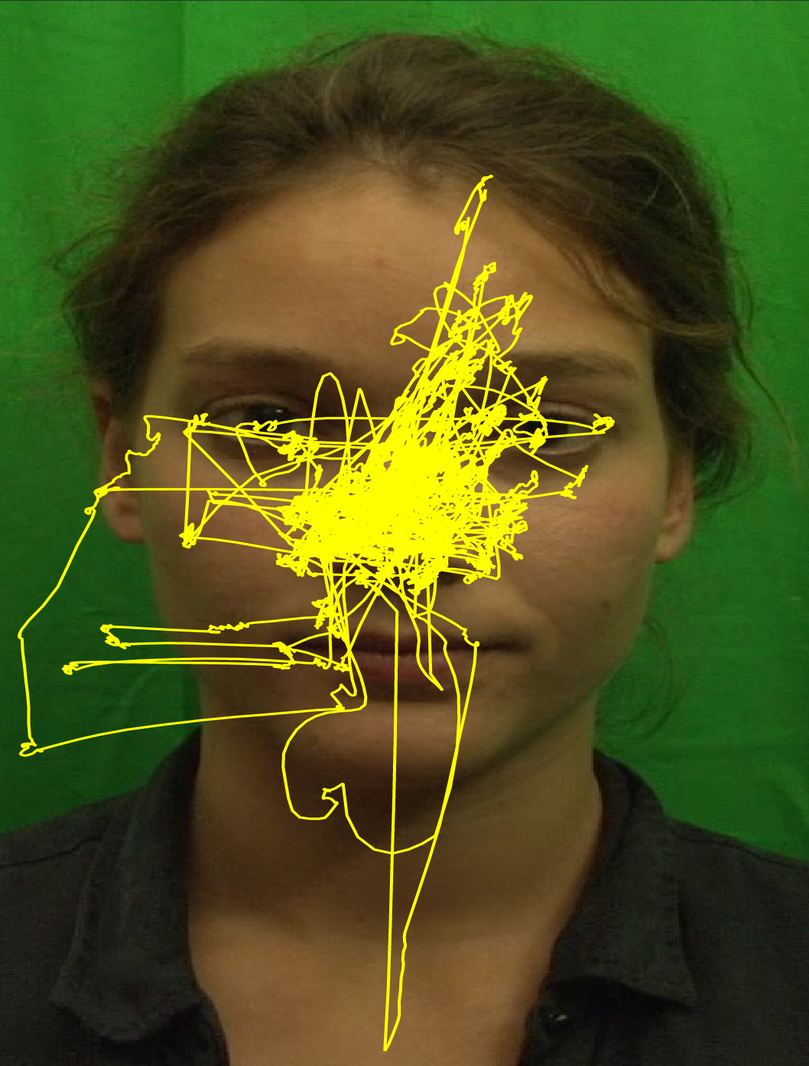}}
		\caption{Male participant} 
		\label{F11} 
	\end{subfigure}
	\caption{Example stimulus with gaze trajectories of a female and a male partici-
		pant.}
	\label{datasetsStimuli1} 
\end{figure}

The minimum recording duration for each participant is two minutes. We capped all recordings to two minutes (a few recordings were slightly longer). 
During the dataset recording each participant had 32 individual trials.
The participants looked at multiple images of a single actor gazing towards them for varying durations (0.1 s to 10.3 s) in all the 32 different trials.
This two minute trajectory is the concatenation of all the 32 trails of each participant. We always merge the last point of the first 
trial with the first point of the next trial.
\autoref{datasetsStimuli1} shows an example stimulus and the trajectories of one female and one male participant.

\subsection{Data Processing and Segmentation}

There is noise in the raw data. To reduce this, a Savitzky-Golay filter~\cite{press1990savitzky, schafer2011savitzky} is used with polynomial order of 6 and a frame size of 15.  
We use the same parameters for the Savitzky-Golay filter as used in~\cite{george2016score, schroder2020robustness}.

For several prediction tasks, segmentation has been shown to be beneficial~\cite{george2016score, sargezeh2019gender, al2020gender}. Hence, we segment the eye movement trajectories into fixations and saccades. 
We use the version of the \emph{identification by velocity threshold}~(I-VT) algorithm used in~\cite{george2016score} and~\cite{al2020gender} to segment eye movements into fixations and saccades.
It uses two parameters: the \emph{velocity threshold}~(VT) and the \emph{minimum fixation duration}~(MFD). 
Any segment of the trajectory for which the velocity is below the~VT, and the length is longer than the MFD is classified as  a fixation; all other segments are classified as saccades.
The standard MFD value of 100~ms is chosen and changing this value (e.g. to 0~ms or 200~ms) has no impact on the accuracy in our study. 
We choose a VT value of 20 degrees/second 
for the dataset. 
By selecting the VT parameter, it is ensured that each participant has a non-zero number of fixations which is around the mean of the maximum number of fixations over all participants.
In previous experiments of~\cite{mahdie2021extensive}, this way of choosing the VT has been shown to 
give the best accuracies of certain classification tasks.

\subsection{Feature Extraction}
\label{sec_feature_extraction}

In the work of~\cite{al2020gender} they show that, contrary to previous research, gender prediction is possible in prepubsecent children with 64\,\% of accuracy using machine learning. As a result of this promising approach, we calculate the sixty-seven features mentioned in~\cite{al2020gender}. 

The work of~\cite{komogortsev2010biometric} mentions that a rich amount of information is present in saccades about the dynamics of the oculomotor plant. Hence, we extract the saccade amplitude and saccadic ratio. 
The temporal properties are leveraged by using the distance and the angle with the previous fixation and saccade as features.  Also, global features like duration, dispersion, and path length are computed.
The derivatives, such as velocity and acceleration are computed using the “forward difference method”.
Earlier work~\cite{harwood2008optimally} suggests that different areas of the brain generate fixations and saccades in horizontal and vertical directions. Therefore, we generate versions in \emph{x} and \emph{y} directions of our features.
Various statistical features (M3S2K) mean, median, maximum, SD (standard deviation), skewness, and kurtosis of the trajectories are computed for velocities and accelerations.

We pick the top eight features (see~\autoref{Featuress}) in order of their ANOVA scores~\cite{de2012comparison, weerahandi1995anova}.
We only consider a maximum of eight features since the accuracy started to drop when we increased the number of features beyond eight.

\begin{table*}[!h]
	\centering
	\caption{Top features ranked using ANOVA scores.}
	\label{Featuress}
	\begingroup
	\setlength{\tabcolsep}{6pt} 
	\renewcommand{\arraystretch}{1} 
	\scalebox{1}{
		\begin{tabular}{ll}
			\toprule
			Fixation features&\begin{tabular}{@{}r@{}}Saccade Features\end{tabular} \\
			\midrule		
			1) Maximum angular velocity & 1) Saccade ratio   \\
			2) SD angular velocity & 2) Mean angular velocity   \\
			3) Minimum angular acceleration  & 3) Mean angular acceleration \\
			4) Mean angular velocity  & 4) SD angular velocity \\
			\bottomrule
	\end{tabular}}
	\endgroup
\end{table*}

\subsection{Machine Learning Classifiers and Performance Metrics}
\label{ML_classifiers}

According to our study on gender prediction, the following classifiers are the most effective:
\begin{itemize}[noitemsep]
	\item Logistic Regression (LogReg)~\cite{bishop2006pattern_chap4}
	\item Random Forest (RF)~\cite{breiman2001random, breiman1996bagging}
\end{itemize} 
We also tried Support Vector Machines (SVM)~\cite{cortes1995support}, Radial Basis Function Networks (RBFN)~\cite{broomhead1988radial, george2016score}, and Naïve Bayes (NB)~\cite{bayes1968naive}, but their results were worse than LogReg and RF.
During the experiments, the ratio of male to female is always balanced. 
We perform hyper-parameter tuning for RF using grid search.

For every experiment, we perform cross-validation using 50 runs. 
Here, cross-validation refers to selecting training and test sets for the prediction randomly for each new run.
In our work, we use the ratio of 80:20 meaning we use 80\,\% participants for training and the remaining 20\,\% for testing.

Our performance metric is the accuracy.
Together with each accuracy, we also report the 
\emph{standard error of the mean}~(SEM),
$\sigma_\mu = \frac{\sigma}{\sqrt{n}}$,
where $\sigma$ is the \emph{standard deviation}~(SD) and $n$ is the number of runs.
This is done using the notation x $\pm$ y\,\%, where x is the accuracy (in percentage points) and y is the SEM. Section~\ref{HELLOO} is the only section in which we also report the
standard deviation (SD).

\section{Gender Prediction Experiments}
\label{sec_Experiments}
\label{Gender_prediction_GOF_data} 
In this section, we describe our experiements for gender prediction. 
We chose an equal number of fixation and saccade features since we have two classifiers, one for fixations and one for saccades.
We perform the experiments using an incremental number of features (see~\autoref{Featuress}) from two to eight and show the achieved accuracies with different classifiers in~\autoref{Accuracies}.
We observe that both classifiers peak with \emph{two} features (one ``fixation'' and one ``saccade'' feature). 
In our study we call these two top features as the ``pipeline features''.
The LogReg classifier gives us the best accuracy of 63.7~$\pm$~0.60\,\% followed by the RF classifier with an accuracy of 63.4~$\pm$~0.70\,\%.

\begin{table*}[!h]
	\centering
	\caption{Accuracies with SEM using top two to eight features over 50 runs.}
	\label{Accuracies}
	\begingroup
	\setlength{\tabcolsep}{6pt} 
	\scalebox{1}{
		\begin{tabular}{ccc}
			\toprule
			No. of features&\begin{tabular}{@{}r@{}}LogReg\end{tabular} &\begin{tabular}{@{}r@{}}RF\end{tabular}  \\
			\midrule		
			2 (1 Sac 1 Fix) & 63.7 $\pm$ 0.60\,\% &63.4 $\pm$ 0.70\,\%  \\
			4 (2 Sac 2 Fix) & 61.8 $\pm$ 0.70\,\% &62.8 $\pm$ 0.80\,\%  \\
			6 (3 Sac 3 Fix) & 61.3 $\pm$ 0.70\,\% &62.6 $\pm$ 0.80\,\%  \\
			8 (4 Sac 4 Fix) & 59.3 $\pm$ 0.80\,\% &62.8 $\pm$ 0.80\,\%  \\
			\bottomrule
	\end{tabular}}
	\endgroup
\end{table*}

\subsection{Optimizing Weights for the Fixation and Saccade Classifiers}

The Nelder-Mead Method~\cite{powell1964efficient} is used to find the optimal weights for the fixation and saccade classifiers. In a survey of black box optimization methods~\cite{hansen2010comparing} for low-dimensional problems, it outperformed the other methods.~\autoref{NBmethod} presents the accuracy and weights for fixations and saccades using equal weights as well as weight-optimized approaches. 
According to the research of~\cite{phdthesis}, the fixation classifier is more important for gender prediction and they found this by optimizing the weights manually. Our study also revealed the same  with the fixation classifier being more important for this task.

The weights increase the accuarcy by 1.5\,\% (see~\autoref{NBmethod}) for the best performing experiment using two features for the LogReg classifier and by 0.6\,\% for the RF classifier.

\begin{table}[!h]
	\centering
	\caption{Accuracies with SEM using Nelder-Mead method over 50 runs.}
	\label{NBmethod}
	\begingroup
	\setlength{\tabcolsep}{6pt} 
	\scalebox{1}{ 
	\begin{tabular}{ccccc}
		\toprule
		\begin{tabular}{@{}c@{}}No. of features\end{tabular} & \begin{tabular}{@{}c@{}}LogReg weights\\ Sac/Fix\end{tabular} & \begin{tabular}{@{}c@{}}LogReg \end{tabular}& \begin{tabular}{@{}c@{}}RF weights\\ Sac/Fix\end{tabular} & \begin{tabular}{@{}c@{}}RF \end{tabular} \\
		\midrule
		 two (1 Sac 1 Fix) & 0.5/0.5 & 63.7 $\pm$ 0.60\,\% & 0.5/0.5 & 63.4 $\pm$ 0.70\,\%\\
		 two (1 Sac 1 Fix) & 0.304/0.696 &65.2 $\pm$ 0.80\,\% & 0.454/0.546 &64.0 $\pm$ 0.70\,\%\\
		\bottomrule
	\end{tabular}}
	\endgroup
\end{table}

\section{Comparison with State-of-the-Art}
\label{HELLOO}
Only in this section we use standard deviation (SD) with the accuracy as our metric.
To the best of our knowledge, the study of~\cite{sargezeh2019gender} has the highest accuracy which uses machine learning classifiers to predict gender using eye movements. They report an accuracy of 70\,\% with an SD of 13.22\,\% using 5 runs. The demographics of the study is as follows: 20 females and 25 males aged 25--34 years. Indoor images are used as stimuli from the Change Blindness database. An Eyelink 1000 plus eye-tracker is used at 1000 Hz. 

In order to compare our pipeline features with the state-of-the-art features we perform a similar experiment (SOTA) using the same number of participants and also using 5 runs. We use two types of features for the experiment. We use our pipeline features and the features of~\cite{sargezeh2019gender}. 
The state-of-the-art features include six features in total. These features are fixation duration, spatial density feature, RFDSD (ratio fixation duration to saccade duration), number of saccades, saccade amplitude, and path length. 
The results are shown in~\autoref{SOTA-NElder}. For the pipeline features, we achieve the best accuracy of 72.5\,\% with an SD of 9.35\,\% using the RF classifier.
And using state-of-the-art features we achieve an accuracy of 55.0\,\% with an SD of 11.00\,\% with the RF classifier.

\begin{table}[!h]
	\centering
	\caption{Accuracies with SEM using pipeline and state-of-the-art features over 5 runs.}
	\label{SOTA-NElder}
	\begingroup
	\setlength{\tabcolsep}{6pt} 
	\scalebox{1}{ 
	\begin{tabular}{cccc}
		\toprule
		\begin{tabular}{@{}c@{}}Feature\\type\end{tabular} & \begin{tabular}{@{}c@{}}No. of \\ features\end{tabular}  & \begin{tabular}{@{}c@{}}LogReg \end{tabular}&  \begin{tabular}{@{}c@{}}RF \end{tabular} \\
		\midrule
		pipeline &two  &  62.5\,\% with an SD of 13.69\,\%  & 72.5\,\% with an SD of 9.35\,\%\\
		\midrule
		SOTA &six  &  52.5\,\% with an SD of 4.36\,\,\%  & 55.0\,\% with an SD of 11.00\,\%\\
		\bottomrule
	\end{tabular}}
	\endgroup
\end{table}

We then optimize the weights for the best accuracy experiment using the Nelder-Mead Method and we find it increases the accuracy by 5\,\%~(see~\autoref{HI-NElder}).

\vspace{20pt}

\begin{table}[!h]
	\centering
	\caption{Accuracies with SEM using Nelder-Mead method over 5 runs.}
	\label{HI-NElder}
	\begingroup
	\setlength{\tabcolsep}{6pt} 
	\scalebox{1}{ 
		\begin{tabular}{cccc}
			\toprule
			\begin{tabular}{@{}c@{}}Feature\\type\end{tabular} & \begin{tabular}{@{}c@{}}No. of \\ features\end{tabular} & \begin{tabular}{@{}c@{}}RF weights\\ Sac/Fix\end{tabular} & \begin{tabular}{@{}c@{}}RF \end{tabular} \\
			\midrule
			pipeline &two   & 0.5/0.5 & 72.5\,\% with an SD of 9.35\,\%\\
			pipeline &two   & 0.219/0.781 &77.5\,\% with an SD of 4.00\,\%\\
			\bottomrule
	\end{tabular}}
	\endgroup
\end{table}

\subsection{Relation between SD and Number of Users}
\begin{figure}[!h] 
	\centering
	
	\includegraphics[width=0.95\textwidth]{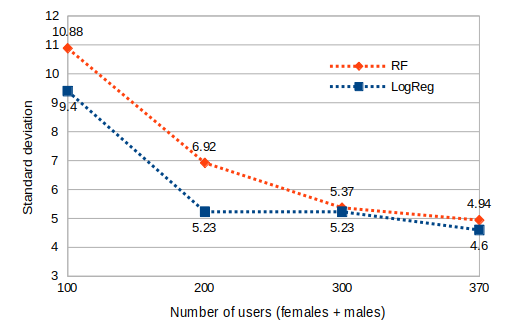} 
	\caption{Standard deviation (SD) for LogReg and RF classifier against number of users with two features and over 50 runs.}
	\label{plot_2} 
\end{figure}

Because we are using SD in this subsection, we wanted to know if there is a correlation between SD and the number of users. 
The \autoref{plot_2} illustrates how, by increasing the number of users, the SD approaches a value at around 4.10\,\%.
According to our experiments, the standard deviation is lower in an environment with more users, and it must approach a constant value as more users are added.

On the other hand, we found that when the number of runs is increased (200 runs), the SEM approaches a value of 0.31\,\% with accuracy of 63.9\,\% using the LogReg classifier. 
This means that in 95.4\,\% of repeating our 50 runs experiment, the resulting mean accuracy will be within the range of (63.3\,\%, 64.5\,\%). 
This is a known statistical fact~\cite{lee2015standard}.
We consider these numbers as stable.

\section{Statistics and Quantitative Observations} 
\label{Statistics}
In this section we discuss some statistics and  quantitative differences between female and male eye movements based on the existing literature~\cite{li2018biometric, sargezeh2019gender, al2020gender}.

We want to verify whether the statistics of the data as mentioned in~\cite{coutrot2016face} are the same. 
They found saccade amplitudes smaller and fixation duration longer for males than for females. 
With our findings (see~\autoref{222}) we confirm that the saccade amplitudes are smaller  and total number of fixations is larger for males. We also find that the path lengths are longer for females. 
This indicates that females are more explorative than males.

\begin{table*}[!h]
	\centering
	\caption{Mean values for females and males.}
	\label{222}
	\setlength{\tabcolsep}{8pt}
	\scalebox{1}{
		\begin{tabular}{ccccc}
			\toprule
			Gender & Path length      & Saccade amplitude & Left eye     & Right eye         \\ \midrule
			Males  & 990.6 $\pm$ 1605.1   & 308.2 $\pm$ 1567.0      & 66.4 $\pm$ 5.2   & 96.2 $\pm$ 8.3  \\
			Females  & 2996.6 $\pm$ 14748.8 & 2064.5 $\pm$ 14741.0  & 89.5 $\pm$ 5.6   & 65.7 $\pm$ 7.2   \\ \midrule

	\end{tabular}}
\end{table*}

The mean number of fixations (in percentage points) in eye region for females and males is shown in~\autoref{222}.
It can be seen that females have more fixations on the left eye of the stimuli. Males on the other hand make more fixations on the right eye.
Previous research of this kind shows very strong left eye bias during the first 250 ms of exploration in women~\cite{leonards2005idiosyncratic}. 
The study of~\cite{sammaknejad2017gender} found that female participants focused their attention more on the left eye of the stimulus than male participants.
It is observed by the work of~\cite{saether2009anchoring} that females have a greater attention to the eye region compared to males. 
All these previous works back our quantitative findings that women have stronger left eye bias compared to males.

\section{Conclusion}
\label{Conclusion}

In this work, we present the first stable results for gender prediction using eye movements.
We trained the data on two classifiers (one fixation and one saccade) and we weighted those classifier probabilities using the Nelder-Mead method.
We were able to achieve accuracies of upto 65.2\,\% using 50 runs and a maximum of two features.
We also increase the state-of-the-art accuracy by 7.5\,\% using our set of features.
Additionally, given the nature of the eye movements, we confirm the left eye bias and explorative behaviour in females in comparison to males.

\section*{Acknowledgement}
We would like to thank Shubham Ajay Soukhiya for his contribution towards this work.
\bibliographystyle{splncs04}
\bibliography{genderref}[]
\end{document}